\pdfoutput=1

\documentclass[11pt]{article}

\usepackage[preprint]{acl}

\usepackage{times}
\usepackage{latexsym}

\usepackage[T1]{fontenc}

\usepackage[utf8]{inputenc}

\usepackage{microtype}

\usepackage{inconsolata}

\usepackage{graphicx}
\usepackage{tikz}
\usetikzlibrary{shapes.geometric, positioning, fit}

\usepackage{listings}

\usepackage{pifont}
\usepackage{stackengine}
\usepackage{multirow}
\usepackage{comment}
\usepackage{cuted,tcolorbox,lipsum} 
\usepackage{xspace}
\usepackage{booktabs}
\usepackage{float}
\usepackage{placeins}
\usepackage{afterpage}
\usepackage{enumitem}
\usepackage{booktabs,siunitx}
\usepackage{colortbl}  
\usepackage{xcolor}    
\usepackage{tabularx} 

\usepackage{hyperref}
\usepackage{soul}
\usepackage{amsmath}
\DeclareMathOperator*{\argmin}{argmin}

\newcommand{\framework}{PAARS\xspace}
\newcommand{\frameworkex}{Persona Aligned Agentic Retail Shoppers (PAARS)\xspace}

\title{\framework: Persona Aligned Agentic Retail Shoppers}

\author{Saab Mansour, Leonardo Perelli, Lorenzo Mainetti, George Davidson, Stefano D'Amato \\
Amazon\\
\{saabm, lperelli, loremain, geodavix, stedama\}@amazon.com
}

\usepackage{xcolor}

\newif\ifsectionenabled
\sectionenabledfalse

\usepackage{setspace}

\begin{document}
\maketitle
\begin{abstract}
In e-commerce, behavioral data is collected for decision making which can be costly and slow. Simulation with LLM powered agents is emerging as a promising alternative for representing human population behavior. However, LLMs are known to exhibit certain biases, such as brand bias, review rating bias and limited representation of certain groups in the population, hence they need to be carefully benchmarked and aligned to user behavior. Ultimately, our goal is to synthesise an agent population and verify that it collectively approximates a real sample of humans. To this end, we propose a framework that: (i) creates synthetic shopping agents by automatically mining personas from anonymised historical shopping data, (ii) equips agents with retail-specific tools to synthesise shopping sessions and (iii) introduces a novel alignment suite measuring distributional differences between humans and shopping agents at the group (i.e. population) level rather than the traditional ``individual'' level. Experimental results demonstrate that using personas improves performance on the alignment suite, though a gap remains to human behaviour. We showcase an initial application of our framework for automated agentic A/B testing and compare the findings to human results. Finally, we discuss applications, limitations and challenges setting the stage for impactful future work.
\end{abstract}

\section{Introduction}
Agents, which are LLMs instructed with a persona (aka role) and have access to tools, are emerging as a powerful paradigm for simulation, surveying and data augmentation. For example, in the domain of election surveying, \cite{argyle2023out} find that GPT3 can emulate a wide variety (they coin the term "algorithmic fidelity") of political human traits with high correlation to human judgement, thus concluding that LLMs encapsulate and can simulate many distributions (when prompted appropriately) rather than a single generic one. \cite{gao2024large} survey the landscape of LLM based agents for simulation, including use cases in the physical domain (e.g. urban environments, ecological and supply chain dynamics) and social or hybrid domain (marketing, economy, opinion, political and market dynamics).

Previous work has primarily focused on using agents to improve performance through collaboration (e.g. \cite{hong2024metagpt}) or for social psychology simulations. However, there has been limited exploration of real-world applications that can benefit from agent-based simulation, particularly in the retail domain. 
LLMs are known to exhibit certain biases that need careful consideration, such as brand bias \cite{kamruzzaman-etal-2024-global}, positive rating bias \cite{yoon-etal-2024-evaluating}, and limited representation of certain groups in the population. These biases can impact the reliability of agent-based simulations and need to be carefully benchmarked and aligned to user behavior.

\begin{figure*}[!t]
  \centering
  \includegraphics[width=1\textwidth]{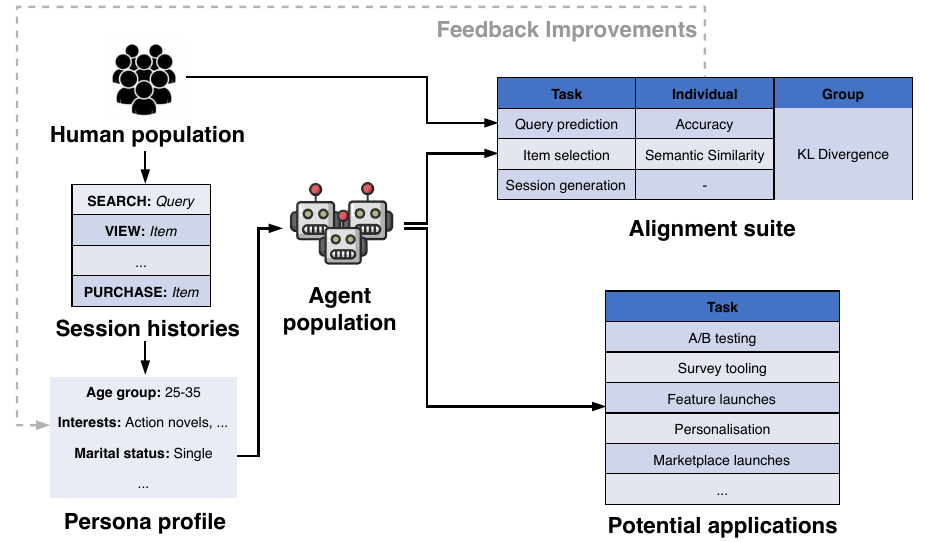} 
  \caption{The \framework framework: we synthesize personas from anonymised human shoppers sessions, generate shopping sessions by powering LLM based agents with personas and retail tools, and measure individual and group alignments to ensure reliability of the persona powered agents. Our framework sets the stage for further persona and session generation improvements, with impactful applications in retail and other domains such as agentic A/B testing and surveying tools.
  }
  \label{fig:framework}
\end{figure*}

In this work, we propose \framework, an LLM-powered agentic framework specifically designed for simulating human shoppers. 
An overview of the framework is shown in Figure~\ref{fig:framework}.
In contrast to previous studies that rely purely on historical behavioral data, we introduce a persona-driven approach where we induce personas from real and anonymised shopping sessions using them to power LLM agents equipped with retail-specific tools (e.g., search, view , purchase, etc). 
The use of induced personas offers several key advantages: 1) it allows measuring algorithmic fidelity across different customer segments, 2) it enables fine-grained population sub-groups experimentation, including potentially underrepresented ones, 3) it mitigates privacy concerns as the personas are synthetic. 

To ensure that our agents behave in a manner indistinguishable from human shoppers, we introduce a behavioral alignment suite. As we care about overall trends for applications such as A/B testing, we focus on distributional differences between human and agentic shoppers behavior at the group (population sample) level. For completeness, we also include traditional ``individual'' alignment tasks. Prior work in LLM-based recommendation and behavioral modeling has primarily focused on individual alignment — measuring how well an agent replicates a specific user's past behavior (e.g., next-purchase prediction). In contrast, our group alignment methodology evaluates whether aggregated shopping sessions of a population of agentic shoppers captures the same distributional properties as human shoppers. This distinction is critical for downstream applications: for example, surveying or A/B testing tools require group-level similarity but do not need each synthetic agent to mimic real individuals accurately.

To summarize the contributions of our work:
\begin{itemize}[noitemsep,topsep=0pt]
    \item Persona powered simulation framework:  we induce personas from historical shopping behavior and equip agents with those personas and retail tools to simulate shopping sessions and measure alignment.
    \item Alignment suite at the group level:  we design a reflective test suite of shopper behavior, proposing a novel formalisation of group alignment aimed at capturing broader population dynamics.
    \item Experimental results: we show that using our personas improves alignment benchmarks though a gap remains to human characteristics, in addition to an initial application of \framework for agentic A/B testing.
\end{itemize}

\section{Related Work}
LLM-powered agents have been widely used to improve performance on complex tasks using multi-agent collaboration. Examples including \citep{hong2024metagpt, qian-etal-2024-chatdev} introduce a structured framework where agents assume distinct roles (e.g., project manager, software architect, engineer, QA) and follow Standard Operating Procedures (SOPs), significantly enhancing coding performance. Similarly, \cite{chen2024agentverse} proposes a framework for breaking down complex tasks into multi-agent collaboration, they extend over previous work by introducing simulation capabilities which is applied to Minecraft game playing showing volunteer, conformity or destructive behaviors. Previous work utilizing agents simulation also focused on studying social behaviors. \cite{aher2023using} replicate human subject studies using LLM agents, identifying distortions in simulating certain behaviors, particularly in education and arts. \cite{park2023generative} propose Generative Agents, an interactive sandbox environment where LLM-driven entities exhibit emergent collaborative skills, such as knowledge diffusion. Further, \cite{zhang-etal-2024-exploring} explores collaboration mechanisms in LLMs from a social psychology perspective, demonstrating how agent personality traits (e.g., confidence, debate, ease of agreement) influence multi-agent reasoning with improvements on the MMLU benchmark. \cite{argyle2023out} introduce the concept of algorithmic fidelity, showing that GPT-3 can emulate various human subgroups, enabling simulations of political traits. While these works focus on social sciences, their application to real-world human behavior modeling, especially in retail, is largely unexplored. Our work extends simulations paradigms by leveraging personas induced from historical behavior to create LLM agents that simulate shoppers decision making.

Ensuring LLM-powered agents behavior aligns to humans is critical for simulation applications. 
To this end, prediction tasks are used in the literature to expose potential biases of LLMs such as brand bias \cite{kamruzzaman-etal-2024-global} or (positive) rating bias \cite{yoon-etal-2024-evaluating}.
Previous work for alignment focuses on emulating user preferences across a diverse set of users. The users are not distinguished by personal traits or preferences. Our proposed framework induces personas from historical behavior data. Interestingly, obtaining unique personas for different users allows us to measure alignment in two ways: 1) Individual alignment - how well an agent replicates the actions of a specific user 2) Group alignment - whether a population of simulated agents exhibits similar distribution to human shoppers. 

Closely related to our work is \cite{yoon-etal-2024-evaluating}, who propose a behavioral alignment suite for LLM-based movie recommendation tasks, including item selection, reviewing, and search query generation.
In the retail domain, \cite{wang-etal-2024-recmind} leverage LLMs for product recommendation, achieving comparable performance to fine-tuned models but relying on historical behavior data only. 
Both works do not explore persona driven modeling. Additionally, given access to historical A/B tests and human sessions, we can directly correlate the tendencies of agentic A/B tests with the human ones.

Overall, our work differentiates itself by integrating persona induction from real shopping data, enabling the creation of agents that simulate shopping sessions reflective of both individual and group behaviors. This methodology not only improves the alignment suite metrics, but also provides a framework for persona powered simulation enabling future impactful applications such as agent based A/B testing and surveying.

\section{Persona Mining}
\label{sec:persona_mining}
The core of \framework framework lies in the \textit{persona mining} methodology, which consists of two subsequent prompting steps.
\begin{enumerate}
    \item \textbf{Consumer profile}: the LLM is provided with a shopping history comprising search, view, and purchase actions from a real customer and is tasked with synthesizing a set of generic user information, such as age range, marital status and income. The shopping history is composed by daily sessions for the last 6 months concatenated with the purchase history older than 6 months.
    \item \textbf{Shopping preferences}: the LLM is prompted one more time with the synthetic consumer profile and the shopping history. The goal is to infer a set of shopping preferences for the user, including factors such as price sensitivity, value perception, reliance on reviews, and brand reputation considerations.
\end{enumerate}

The complete persona is eventually composed by: a consumer profile, shopping preferences, and the real shopping history itself. Additionally, the reasoning produced by chain-of-thought prompting is also included in the persona.
The mined personas will represent the role the LLM needs to impersonate while generating responses across the various tasks we define. 
An illustrative shopping session and induced persona are shown in Appendix~\ref{appendix:persona}.

\section{Alignment Evaluation Suite}
\label{sec:alignment}
In this work, we propose an alignment suite aimed at evaluating the degree to which an agent population approximates a human population in the domain of e-commerce tasks.
The alignment suite evaluates alignment at two different levels, namely (i) individual level and (ii) group level. 

\subsection{Individual and Group alignment}
Let us define a population of human shoppers $\mathbf{H} = \{h_i\}_{i=1}^{n}$. 
Our goal is to create a synthetic population of shopper agents $\mathbf{A} = \{a_i\}_{i=1}^{n}$ that mimics $\mathbf{H}$. 
To evaluate how well the synthetic population mimics the human population, most works in the literature focus on what we refer to as individual alignment~\cite{jannach2010recommender}. 
This measures how well each $a_i$ approximates its corresponding $h_i$ across different tasks. 
Below, we expand on this notion and formalise the concept of group alignment, which measures how well the whole population $\mathbf{A}$ approximates $\mathbf{H}$. 
Note that while we set the cardinality of both groups to be the same such that individual metrics can be defined, this is not required for group metrics.
To formalise individual and group alignment for a given task $\mathbf{T}$, let us define the outputs of the human shoppers $\mathbf{O_H} = \{o_{h,i} | o_{h,i} = h_i(\mathbf{T})\}_{i=1}^{n}$ and outputs of the agentic shoppers $\mathbf{O_A} = \{o_{a,i } | o_{a,i } = a_i(\mathbf{T})\}_{i=1}^{n}$. 

\paragraph{}

For \textbf{individual alignment}, the metrics directly compare an agent to its human counterpart. 
If $\mathbf{f_{agg}}$ is an aggregation function and $\mathbf{f_{comp}}$ is a comparison function, individual metrics $\mathcal{M}_{individual}$ can generally be computed as:
\begin{equation}    
\mathcal{M}_{individual} = \mathbf{f_{agg}} ( \{\mathbf{f_{comp}}(o_{a,i}, o_{h,i}) \}_{i=1}^{n})
\end{equation}
For example, for classification accuracy $\mathbf{f_{agg}}$ is the average function while $\mathbf{f_{comp}}$ is the equality function with binary output. 

\paragraph{}

For \textbf{group alignment}, let us define $\Phi$ as some measure of distributional dissimilarity between two distributions. 
Then, we say that $\mathbf{A}$ is group-aligned to $\mathbf{H}$ on $\mathbf{T}$ if $\mathbf{O_H}$ approximates $\mathbf{O_A}$ in distributional sense, and we measure this through group metrics $\mathcal{M}_{group}$ computed as:
\begin{equation}    
\mathcal{M}_{group} =\Phi(\mathbf{O_H}, \mathbf{O_A})
\end{equation}
In this work, we set $\Phi$ to be the Kullback-Leibler divergence. 
To highlight the difference between individual and group alignment metrics, one can notice that the 1-dimensional KL divergence can also be expressed as:
\begin{equation}
\mathcal{M}_{group} = \mathbf{f_{comp}} ( \mathbf{f_{agg}}(\{o_{a,i}\}_{i=1}^n), \mathbf{f_{agg}}(\{o_{h,i}\}_{i=1}^n))
\end{equation}
where $\mathbf{f_{agg}}$ is a binning function while $\mathbf{f_{comp}}$ is the sum of relative entropies of the bins. 
Given two candidate shopper agent populations $\mathbf{A_1}$ and $\mathbf{A_2}$, $\mathbf{A_1}$ has better group alignment to $\mathbf{H}$ if $\Phi(\mathbf{O_H}, \mathbf{O_{A_1}}) < \Phi(\mathbf{O_H}, \mathbf{O_{A_2}})$. 
Moreover, our framework adopts LLMs as backbones for the population of agents $\mathbf{A}$. Since we aim to build a large population of agents and fine-tuning each agent is very expensive, we use the same set of weights for all agents and equip each agent $a_i$ with a specific prompt $p_i$ , which we refer to as “persona”. We use $a_i \langle  p_i \rangle$ to denote the agent $a_i$ induced by $p_i$.
The agent population induced by a set of personas $\mathbf{P} = \{p_i\}_{i=1}^{n}$ is denoted as $\mathbf{A(P)}$, where:
\begin{equation}
\mathbf{A(P)} = \{a_i \langle  p_i \rangle | p_i \in \mathbf{P}  \}
\end{equation}
Then, given a human population $\mathbf{H}$ and some task $\mathbf{T}$, finding the optimal agent population $\mathbf{A(P)}$ can be formulated as extracting the set of prompts $\mathbf{P^*}$ that minimise distributional dissimilarity:

\begin{equation}
\mathbf{P^*} = \underset{\mathbf{P}} {\argmin} \hspace{0.3cm} D_{\mathrm{KL}}(\mathbf{O_{H}}, \mathbf{O_{A(P)}}),    
\end{equation}

which also includes the case where no persona is provided if $p_i$ is an empty string.
We believe that group alignment is an interesting direction to explore for three reasons, namely: (i) it is a weaker condition compared to individual alignment, which can be challenging to achieve in many cases, (ii) it can be used to compare populations of different sizes, allowing for greater flexibility in the construction of agent populations and (iii) it can predict useful signals due to its aggregate nature. 
For example, consider the case of A/B testing simulation, where the goal is to predict directional results from a human population of shoppers through a surrogate population of agents. 
Then, the A/B test outcome direction can be predicted correctly even with low individual alignment, as long as group alignment is sufficiently high. 
The former is very difficult to achieve due to the complexity of human behaviours when taken as a single entity, but the latter is more tractable when considering larger populations. An example of individual metrics contrasting with group metrics is provided in Appendix \ref{appendix:group_vs_individual}.

\subsection{Computing KL divergence}
Since the outputs of the tasks can be either mono-dimensional (i.e. a histogram of relative frequencies) or multi-dimensional (i.e. embeddings), we report how the KL divergence is computed in the two cases.
In general, the KL divergence for the discrete case can be computed as:
\begin{equation}
D_{\text {KL }}(P \| Q) =\sum_{x} P(x) \log \frac{P(x)}{Q(x)}
\end{equation}
where $P$ and $Q$ are two probability distributions.
For the mono-dimensional case, we compute $P(x)$ and $Q(x)$ using histograms (relative frequencies) over bins $x$ corresponding to discrete categories or intervals. 
For the multi-dimensional case instead we approximate the KL divergence through the following Monte Carlo estimator:
\begin{equation}
D_{\mathrm{KL}}(P \| Q) \approx \frac{1}{N} \sum_{i=1}^N\left[\log P\left(x_i\right)-\log Q\left(x_i\right)\right]
\end{equation}
with samples $x_i$ taken from $P$, while $P(x_i)$ and $Q(x_i)$ are estimated with a kernel density estimator (KDE). Details on the binning, histogram calculation and KDE are given in the group alignment sections in the task definitions below.

\subsection{Task definitions}
In the following paragraphs we describe the 4 main tasks that make up the alignment suite.
The alignment suite covers tasks that appear in a simplified but common customer shopping journey, which includes:  (i) performing a search query, (ii) selecting an item to view the details of the product, and (iii) purchasing a product.

\paragraph{Query generation} To estimate the agent's capability of generating search queries, we devise the task of predicting queries given a product, i.e. what is a plausible query that would have led to viewing the given product title. To mine <search query, viewed product> pairs, we collect human shopping sessions that include actions of search query and product viewing with timestamps. We take the first search query in a session and a consecutive viewed item within 60 seconds. We discard the remaining of the session as some of the queries are challenging to guess without the full context. For example, in a session with following sequential behaviors: (i) search:``soccer shoes indoor'', (ii) view: ``<brand> Unisex-Adult <model-name> Indoor Sneaker'', (iii) search: ``<model-name>'', (iv) view: ``<brand> <model-name> Og Mens Shoes'', it is challenging to predict the second query (ie: ``<model-name>``) from the last view only. We collected 3058 <search query, viewed product> pairs under the former conditions to form the test set. 

To measure \textit{individual alignment}, we compute the match frequency between an agent generated query and the ground-truth human query computing the weighted similarity scores while stratifying data based on query complexity (perplexity) levels.
First we compute the embedding of the predicted query and the real human associated query using the all-MiniLM-L6-v2 \citep{reimers-2019-sentence-bert} model; then we adopt the cosine similarity metric between the two embeddings as similarity score.
In our approach, we assess the difficulty of predicting a query by computing the conditional perplexity of a language model given the sequence of associated views' actions. Specifically, we first construct a context by concatenating the viewed item and then appending the predicted query. The text is tokenized and perplexity is calculated using the GPT-2 LM. 

To measure \textit{group alignment}, we estimate a probability distribution over the space of the query embeddings and compute KL divergence between the human population and agent population, with/without persona.
The probability distribution is estimated through the Kernel Density Estimation \cite{kde} algorithm using Gaussian kernels, while the KL divergence is estimated through the multi-dimensional Monte Carlo estimator due to the continuous and high-dimensional nature of the embeddings space. 

\paragraph{Item selection} For \textit{individual alignment}, we provide 4 items to the agents and task them with purchasing one of these 4 items, then we measure how often the agents purchase the same item that was bought by the corresponding human.
The test cases are constructed as follows:
\begin{itemize}[noitemsep,topsep=0pt]
    \item One item is the ground truth purchase that the human bought. 
    \item The other three items are selected from a list of 1000 random items, but based on non overlapping interests among that persona's interests. 
    \item To avoid data contamination, for each test case we remove all occurrences (views and purchases) of all 4 items from the shopping history that the agent is conditioned on.
\end{itemize}
Each item is represented by a product title and a product category, and the LLM is prompted in different setups, for example with/without the persona conditioning. The final dataset is composed of 4600 test cases. We measure the performance on this task by computing the accuracy of the purchases prediction.

For \textit{group alignment}, we provide the agents with a ranked list of items, corresponding to the search results of a query in the human's shopping history. 
We then prompt the agents to select a product that the corresponding human customer would likely choose to view, and we record its ranked position in the search results. 
Finally, we compute the discrete probability distribution over ranks for the selected items and compute KL divergence to compare human distribution against agent distributions, with/without persona. 

\paragraph{Session generation} 
Our setup involves having the agents interact with a simulated environment mimicking the actual retail website experience. 
In this work, the environment is textual-only, meaning we have no product images and no graphical interface.
The agents are equipped with a set of tools that let them interact with the website, namely: (i) Search tool, which performs a query and returns search results, (ii) View tool, which provides detail page information for a specific product (description, bullet points, reviews,...), (iii) Cart tool, which lets agents add/remove items to their cart. 

For this task we only track group level metrics. 
To measure \textit{group alignment}, we track per-session statistics such as \#searches, \#views and \#purchases.
Then, we compute the discrete probability distribution and compute KL divergence to compare human distribution against agent distributions, with/without persona.
The dataset for this task comprises of 2400 sessions for each configuration.

\section{Experimental Results}
\label{section:results}
In the following sections we highlight the experimental results obtained on the previously introduced alignment suite, measuring both individual and group alignment. 
While the proposed framework is agnostic to a specific model choice, we use Anthropic Claude Sonnet 3.0 as the LLM backbone and  refer to it as Base.
Unless otherwise stated, we default to a $temperature=0$ for all experiments.
All prompts used for persona generation and alignment are provided in Appendix \ref{appendix:prompts}.

\begin{figure}[!t]
  \centering
  \includegraphics[width=1\linewidth]{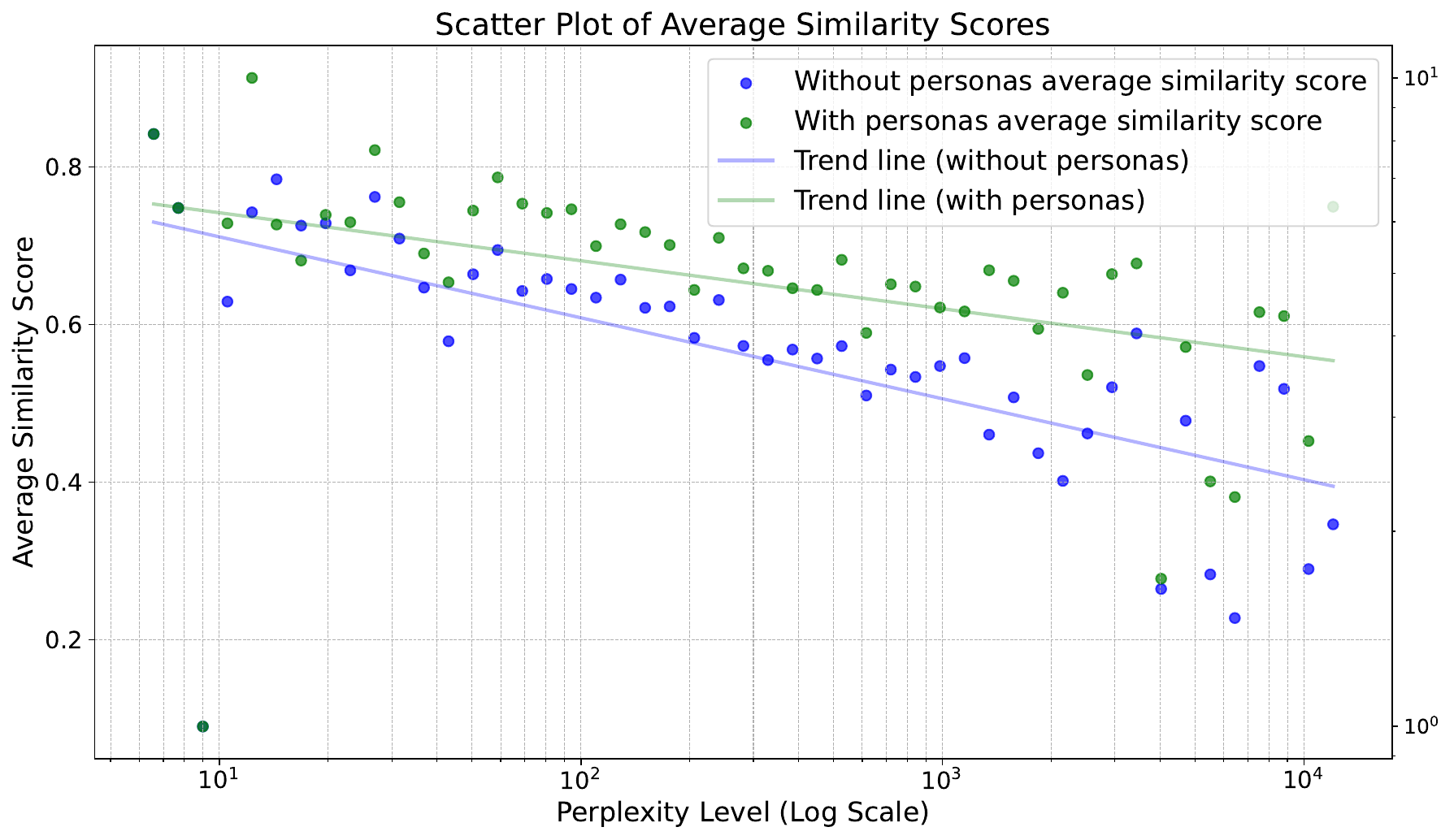} 
  \caption{
  Query generation task: we compare agents with and without personas, by measuring the cosine similarity of the agentic queries against the human ones across different query perplexity levels.
  }
  \label{fig:query_prediction}
\end{figure}

\paragraph{Query generation} For \textit{individual alignment}, we measure the average similarity score between the human queries and the corresponding agent generated ones. We obtain an average similarity score with personas of 0.69 against a score of 0.59 without personas.
We observe that additional conditioning on the persona leads to a 17\% relative improvement.
We plot a graph of similarity scores across increasing levels of human query perplexities in Figure~\ref{fig:query_prediction}. 
We note a correlation between query perplexity increase and query similarity score decrease, which is expected as higher perplexity queries should be more difficult to predict for the agents. 
Nonetheless, the inclusion of personas is beneficial at all levels of perplexity.  
Table \ref{tab:query_qualitative} shows an example of query prediction with and without persona, highlighting better alignment to human queries when conditioning on personas.

\begin{table}[]
\begin{tabular}{lc}
\toprule
\textbf{Method} & \textbf{Query}\\
\midrule
Baseline & knee brace for pain relief\\ 
\hspace{0.2 cm}+ Persona & knee brace for women                                        \\ 
\hline
Human & adjustable knee brace for women \\ 
\bottomrule
\end{tabular}
\caption{Example of query prediction with/without persona.}
\label{tab:query_qualitative}
\end{table}

For \textit{group alignment}, we first use Kernel Density Estimation to estimate a probability distribution over the space of embeddings, which are 384-dimensional.
We use a Gaussian kernel and select a bandwidth of $0.1$. 
Then, we use Monte Carlo estimates with $1000$ samples and repeat the estimation 5 times, averaging the results.
Table \ref{tab:group_alignment_all} shows that agents with persona achieve lower KL divergences from the human distribution compared to agents without persona.
Appendix \ref{appendix:bandwidth} shows that results are consistent for different choices of bandwidth.

\begin{table*}[h]
\centering
\begin{tabular}{lccccc}
\toprule
& \textbf{Query generation} & \textbf{Item selection} & \multicolumn{3}{c}{\textbf{Session generation}} \\ 
\midrule
& \textbf{} & \textbf{} &  \multicolumn{1}{c|}{\# Searches} & \multicolumn{1}{c|}{\# Clicks} & \# Purchases \\ 
&  & & \multicolumn{1}{c|}{} & \multicolumn{1}{c|}{} & \\ 
\hline
Base & 18.81 & 2.40  & \multicolumn{1}{c|}{11.69} & \multicolumn{1}{c|}{11.70}              & 11.68 \\ 
\hspace{0.2cm} + Persona & \textbf{17.51}            & \textbf{1.08}                   & \multicolumn{1}{c|}{\textbf{3.71}}        & \multicolumn{1}{c|}{\textbf{3.72}}      & \textbf{3.68}  \\ 
\bottomrule
\end{tabular}
\caption{Group alignment metrics - KL divergence between human distribution and agent distribution with/without persona for all the tasks in the alignment suite.}
\label{tab:group_alignment_all}
\end{table*}

\paragraph{Item selection} 
Table \ref{tab:item_selection_results} shows the results of the \textit{individual alignment} task. 
We conducted a comparative analysis between the Base model and various persona configurations, first ablating the impact of single persona components in isolation, and then assessing the performance of the complete persona, which is a concatenation of the single components.
The Base model achieves 25.46\% which is close to random guessing. This is expected as it does not have any customer specific information. 
Providing more customer context through personas greatly increases accuracy, achieving an improvement of +6.15\% over the best baseline model (using history only).

\begin{table}[h]
\centering
\begin{tabular}{lc} 
\toprule 
\textbf{Shopping Background} & \textbf{Accuracy (\%)} \\
\midrule 
Base                   & 25.46 \\
\hspace{0.2cm} + Consumer profile & 35.95 \\
\hspace{0.2cm} + Shopping Preferences & 39.01 \\
\hspace{0.2cm} + History & 41.11 \\
\hline
\hspace{0.2cm} + Persona & \textbf{47.26} \\
\bottomrule
\end{tabular}
\caption{Item selection individual alignment - purchase prediction task. The accuracy for using different shopping background levels is provided.} 
\label{tab:item_selection_results}
\end{table}
 
\begin{figure}
    \centering
    \includegraphics[width=1\linewidth]{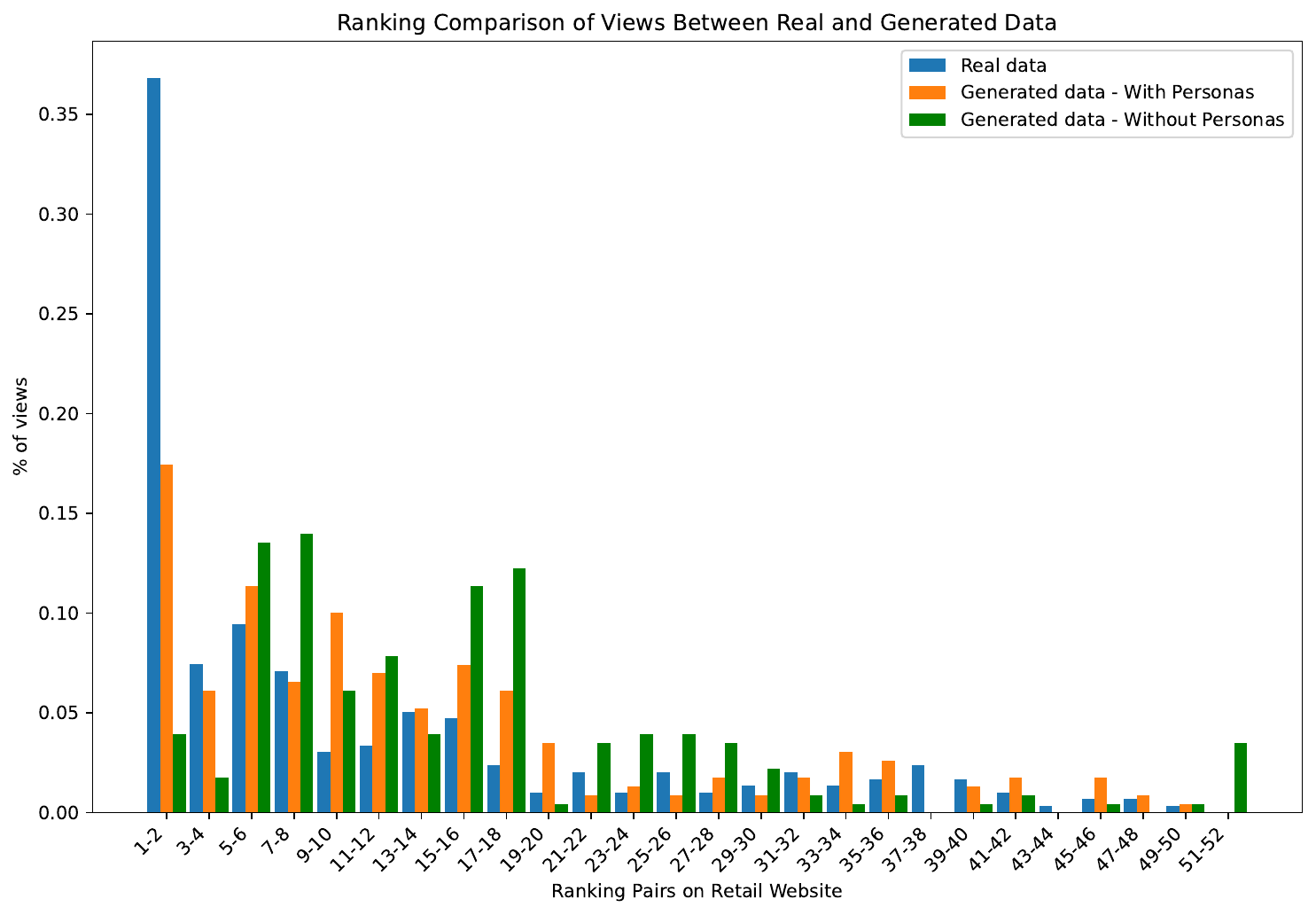}
    \caption{Search rank distribution of viewed items comparing human behavior to agents with/without personas.}
    \label{fig:ranking_distribution}
\end{figure}

For \textit{group alignment}, Table \ref{tab:group_alignment_all} shows that agents with persona achieve better results than agents without in comparison to humans.
This can be explained by observing Figure \ref{fig:ranking_distribution}, which shows the distribution of selected items for the 3 populations. 
While all distributions exhibit a decreasing trend, indicating that higher-ranked items are more frequently viewed, the distribution induced by persona-equipped agents has better alignment with humans.

\paragraph{Session generation} 
For this \textit{group alignment} task we set $temperature = 0.5$ to ensure that agent sessions can evolve differently. 
The results are reported in Table \ref{tab:group_alignment_all}.
As for the previous tasks, persona conditioning lowers the KL divergence. 
Additionally, we compute token-type-ratio (TTR) for both the queries and the titles of products viewed. 
Table \ref{tab:session_generation_ttr} shows that the human population achieves greatest diversity in queries searched and products viewed, with a gap over agent populations.
However, we note that including personas leads to improved diversity.

\begin{table}[h]
\centering
\begin{tabular}{lcc} 
\toprule
\textbf{Method} & \textbf{Query-TTR} & \textbf{Product-TTR} \\
\midrule 
Base & 0.013 & 0.035 \\
\hspace{0.2cm} + Persona & \textbf{0.23} & \textbf{0.66} \\
\hline
Human & 0.38 & 0.97 \\
\bottomrule
\end{tabular}
\caption{TTR metrics for queries searched and products viewed by human and agent populations.} 
\label{tab:session_generation_ttr}
\end{table}

\paragraph{A/B testing simulation} 
To showcase the potential of \framework in complex applications, we conducted an initial limited effort to simulate three real A/B tests. We plan to scale A/B testing simulation in the future to ensure reliability and significance. 
We used the same setup from the \textit{Session generation} task to simulate historical A/B tests carried out on the retail website. 
We target specific A/B tests whose main effects can be simulated in a textual environment, for example change of result ranking algorithms.
We set up two environments, one for Control (C) and one for Treatment (T), where the two differ solely based on their search result and product content such that these match what was viewed by customers during C and T respectively.
For A/B metrics, we track the Sales change between the C and T, correlating results from agents with these from real customers. 
Overall, we measured a directional agreement in Sales change between historical and simulated test for 2 A/B tests out of 3.
On the other hand, we note that the magnitude of Sales change is bigger for the simulated environments (10-30x). We hypothesize that the magnitude increase is a result of the way we define shopping intentions for sessions, where the agents are biased towards purchasing items. Intentions improvements, in addition to scaling the number of personas and A/B tests are left for future work.

\section{Discussion}
Our \frameworkex framework offers a robust and scalable approach to modeling and analyzing human shopping behavior through synthetic persona driven agents. \framework opens up several promising applications while also highlighting important limitations and ethical considerations that need to be addressed.

A general advantage of persona induction is the enhanced explainability of the system. Traditional retail modeling often relies on historical behavioral data without explicit persona modeling, making it difficult to understand why certain purchasing patterns occur. Associating historical behaviour and purchase data with an induced persona allows debugging of simulation and prediction results.

One of the most impactful applications of \framework is simulated A/B testing, offering an offline (i.e. automated, not involving humans) testing environment before deployment. This capability is valuable in accelerating the experimentation cycles and increasing the likelihood of success of human based A/B testing. In addition, \framework facilitates agent-based surveying, where synthetic shoppers can respond to market research queries. User experience research is fundamental to retail feature launches and relies heavily on slow and costly human surveys. In some cases it is even challenging to perform offline testing for UX features, e.g. to introduce a feature of reviews summarization one can measure the quality of the summaries but we can not predict whether humans would prefer to see the feature without resorting to surveys (or human A/B testing). Simulating large-scale consumer analysis without the logistical constraints of traditional human surveys can serve as an effective guardrail to feature launches.

Another key application of \framework lies in targeting underrepresented consumer groups. Traditional behavioral data is often skewed toward majority demographics, leaving gaps in understanding the shopping experiences of marginalized or less-studied communities. By generating synthetic personas that model diverse backgrounds and purchasing behaviors, \framework enables businesses to proactively design more inclusive shopping experiences. This ability to simulate different market segments is particularly useful for new feature and product launches, where companies need to predict how various consumer groups will interact with new offerings.

Although our experimentation is limited to the retail domain, \framework is highly generalizable and can be applied to various fields requiring behavioral forecasting. Simulation is important in many domains including economy and finance (e.g. supply and demand, stock market), public policy and urban planning~\cite{gao2024large}.

\section{Conclusions}
In this work, we introduced \framework, a novel framework to simulate human shoppers behaviour by leveraging persona driven LLM agents alongside an alignment suite at the individual and group levels to ensure reliability. Our experiments show improvements on the alignment tasks when using personas over historical behaviour only. While we demonstrate an initial application for A/B testing, the framework can be further applied to surveying and market research among others. 

Beyond retail, the framework generalizability makes it applicable to a wide range of domains requiring simulations, including financial markets, public policy and urban planning. Given the large potential, \framework opens the door for challenging and impactful research, such as persona and session generation improvements, multilinguality and multimodality, algorithmic fidelity across marketplaces and research into cultural nuances.

\section{Limitations and Future Work}
While \framework offers a promising direction in retail behavior simulation, several challenges remain. Improving the personas and session generation capabilities are important to ensure alignment to human behavior and successful downstream applications. Additionally, the alignment suite can be expanded with more fine-grained testing, such as nuanced cultural evaluation, and additional tasks like navigation or filtering experiences in e-commerce. Furthermore, independent of personas, fundamental testing of LLMs may be necessary, particularly for smaller models, including sensitivity testing to price changes, review ratings, and brand bias. 

Our current experimentation is limited to the text modality, the English language and the US marketplace. Extending to the global marketplace brings new challenges such as testing for cultural nuances and the model fidelity of representing diverse and global populations of shoppers. Many features in retail include visual elements that impact the user journey and decision making. Including the visual modality and leveraging multimodal LLMs is essential for accurate simulation. We hypothesize that due to the large scale behavioural data one can continuously fine-tune and align LLMs for accurate modeling. This brings another challenge of maintaining personas over time. Consumer preferences are dynamic, yet updating synthetic personas in a way that reflects real-world shifts in a cost effective manner remains an open question. 

\section{Ethical Considerations}
As with any LLM-powered system, \framework is susceptible to biases inherent in the models' training data. The under-representation of certain groups in publicly available datasets can lead to reduced model fidelity. 
Specialized research is needed to ensure that the framework remains fair and representative across different demographic segments. 

Even though personas generated by \framework are synthetic, they must be handled with appropriate security and privacy considerations. Methods such as differential privacy can be applied, providing formal guarantees that individual user contributions to the training data are not traceable even if an adversary has access to the model outputs. 

Furthermore, while \framework enhances the efficiency of A/B testing and market research, it cannot fully replace human participation in these processes. Consumer behavior is dynamic and influenced by psychological and social factors that LLM-powered agents may not fully capture. The periodic validation of synthetic personas against real human shoppers is necessary to ensure that simulations remain grounded in actual user experiences. Additionally, LLM-based agents lack the ability to engage with physical products, meaning that some aspects of the shopping experience such as ``trying'' the product cannot be realistically simulated. Hence, we suggest that \framework should be leveraged as a complementary tool that augments human-driven research.

\bibliography{custom}

\newpage
\appendix

\section{Example Persona}
\label{appendix:persona}
As an example we show below an illustrative (inspired by real data) excerpt of a shopping session followed by the induced persona:

\begin{tcolorbox}[colback=gray!10, colframe=gray!50, title= Shopping session]
\small
{\ttfamily
2024-09-10 \\
========== \\
<SEARCH> waterproof hiking shoes - at 10:12 \\
<VIEW> Men's Low height boots - at 10:14 \\
<SEARCH> hiking boots - at 10:35 \\
<VIEW> <Brand1> Waterproof hiking boots - at 10:35 \\
<PURCHASE> <Brand1> Waterproof hiking boots - at 10:42 \\
========== \\
2024-09-12 \\
========== \\
<SEARCH> best solo travel books  - at 14:22 \\
<VIEW> The full guide to solo traveling - Paperback - at 14:33 \\
<PURCHASE> The full guide to solo traveling - Paperback - at 14:50 \\
}
\end{tcolorbox}

The corresponding induced persona:
\begin{tcolorbox}[colback=gray!10, colframe=gray!50, title= Induced persona]
\small
{\ttfamily
Profile: \\
Age Group: 30-45 \\
- Reason: Interest in solo travel and gear \\
Relationship: Single \\
- Reason: Purchases solo travel books \\
Interests: Hiking, camping \\
Shopping Preferences: \\
Brand Reputation: \\
- Prefers <Brand1>, <Brand2>, <Brand3> \\
- Researches best-rated travel books \\
Price Sensitivity: \\
- Willing to invest in durable outdoor gear \\
- Prefers paperbacks over hardcovers \\
Value Perception: \\
- Invests in gear for long trips \\
- Buys books for self-improvement and travel \\
}
\end{tcolorbox}

\section{Group metrics vs Individual metrics}
\label{appendix:group_vs_individual}
Consider the following simplified experimental setting: you want to predict the outcome of a 5 sided dice, with sides ranging from 1 to 5. You consider two systems: system A always predicts 3, while system B always predicts a random number from 1 to 5. 
On individual metrics such as accuracy of prediction or MSE, system A performs equal to or better than system B. However, their group alignment metrics differ substantially, with system B achieving $0$ KL divergence in expectation. 
\begin{table}[h]
\centering
\begin{tabular}{lccc} 
\toprule
\textbf{System} & \textbf{MSE} & \textbf{Accuracy} & \textbf{KL} \\
\midrule 
A & \textbf{1.97} & \textbf{20.6\%} & 10.04 \\
B & 3.96 & \textbf{20.3\%} & \textbf{0.0095} \\
\bottomrule
\end{tabular}
\caption{Individual metrics (MSE, Accuracy) vs Group metrics (KL Divergence) on the random die task.}
\label{tab:group_vs_individual}
\end{table}
This is shown in Table \ref{tab:group_vs_individual}, obtained by simulating 1000 dice tosses.
Now consider this data is used to inform decisions that rely on group alignment. While individual metrics are unable to pick up this nuance and fail to suggest system B being better suited than system A, group alignment metrics capture this. 
Extending this concept from evaluation to actual training objectives, group alignment metrics can produce models which are better suited for the class of tasks we are considering in this work.

\onecolumn
\section{Prompts}
\label{appendix:prompts}
\subsection{Persona mining}
\subsubsection{Consumer profile}
\begin{lstlisting}[language=Python]
"""
The user's data is provided within <user_data></user_data> XML tags. 
The user's data includes sessions data inside <sessions_history> XML tag and all 
other purchases in the <other_purchases> tag:

<user_data>
    <sessions_history>
    {sessions}
    </sessions_history>

    <other_purchases>
    {other_purchases}
    </other_purchases>
</user_data>

Please create a consumer profile that includes the following fields:
- Gender
- Age
- Relationships
- Education
- Industry
- Salary range
- Home ownership
- Parental status
- Interests
    
For each field in the consumer profile, you must provide reasoning behind your 
decision. If you are not certain about a particular field, make your best guess 
based on the available information.
Choose Interests only from the following list: 
    <valid_interests>
    {valid_interests}
    </valid_interests> 
and save them as a list.

Before providing your final output, please think through your analysis using the 
following steps:
1. Carefully review the user's purchase history, view history, and search history.
2. For each consumer field, identify patterns or specific items that could indicate the user's characteristics.
3. Consider how different pieces of information might relate to each other to form a coherent profile.
4. If you're unsure about a field, look for indirect clues that might support a reasonable guess.

After completing your analysis, provide the consumer profile in JSON format. 
Each field should include both the determined value and extensive reasoning 
behind it.

Remember to include all nine required fields in your JSON output, even if you have 
to make a best guess for some of them.

After completing the consumer profile task, provide your output in JSON format with 
two keys. 
Remember to associate the analysis to the key 'analysis' and the consumer profile 
to the key 'consumer_profile'.

Here is an example of the expected JSON output. It is enclosed in the 
<output></output> XML tag.

<output> {example_output} </output>
"""
\end{lstlisting}

\subsubsection{Shopping preferences}
\begin{lstlisting}[language=Python]
"""
You will receive a consumer profile, in the <consumer_profile> XML tag, and history 
data, including sessions in <sessions_history> and all the other purchases not 
included in the sessions in the <other_purchases> tag. All these information are 
enclosed in <user_data></user_data> tag.

<user_data>
    <consumer_profile>
    {consumer_profile}
    </consumer_profile>

    <sessions_history>
    {sessions}
    </sessions_history>

    <other_purchases>
    {other_purchases}
    </other_purchases>
</user_data>

Your goal is to create a persona that describes how this individual might consider 
the following factors while shopping on an online store:
- Products
- Price
- Value
- Product Selection
- Reviews
- Brand Reputation
- Quality

Analyze the provided user data carefully. Look for patterns in their purchase 
history, view history, and search history. Consider how their consumer profile might influence their shopping behavior.

Create a cohesive persona that represents this user's likely approach to online 
shopping. The persona should feel like a real person with distinct preferences 
and behaviors.

In your response, describe how the persona might prioritize each of the seven 
factors listed above. For example, they might prioritize brand reputation over 
price, or value over product selection.

The output must be in JSON format. 

Use "inner_monologue" key to show your reasoning process as you analyze the data 
and form the persona. This will help explain how you arrived at your conclusions.

Present your final persona description in the "persona" key. 
The persona should be written in paragraph form, describing the individual's 
approach to online shopping and how they consider each of the seven factors. 
Be sure to mention the relative importance of each factor to this persona.

Remember, the goal is to create a realistic and specific persona based on the 
provided data, not a generic description. Your persona should reflect the unique 
characteristics and preferences suggested by the user data.
"""
\end{lstlisting}

\subsection{Alignment suite}
\subsubsection{Query generation}
\begin{lstlisting}[language=Python]
"""
Your mission is to analyze given persona characteristics and viewing session data, 
then predict the most likely search queries for each session. This task requires 
careful consideration of the persona's preferences, interests, and behavior 
patterns.
    
Follow these instructions carefully:

1. You will be given a set of persona characteristics inside <persona> tags. Embody this persona for the task. <persona>{persona}</persona>

2. You will be presented with a list of sessions inside <sessions> tags. Within is session there is a list of product viewed. <sessions>{sessions}</sessions>

3. For each session you have to predict the query the user you are embodying has done. Consider the following: 
    a. Analyze the persona characteristics and infer the person's preferences and pickiness level. 
    b. Predict the most probable search query that led to viewing those items. 
    c. Make a decision that best fits the persona's likely preferences and pickiness level that you need to infer.

4. Your final output must be in valid JSON format, containing one key-value pair per session. The key should be the session name (as provided in the input), and the value should be your predicted query for that session.

5. Here's an example of the expected output in JSON format: {example_output}

6. Before making your final predictions, use this section to think through your reasoning:

Persona analysis:
- What are the key characteristics of this persona?
- What preferences and interests can you infer?
- How might these traits influence their search behavior?

Session analysis (for each session):
- What types of products were viewed?
- How do these align with the persona's characteristics?
- What common themes or purposes can you identify among the viewed items?

Query prediction:
- Based on the persona and viewed items, what search terms are most likely?
- How specific or general should the predicted query be?
- Does the predicted query align with the persona's likely language and search style?
- Use these thought processes to inform your final predictions.

7. Important notes:
- Only predict one query for each sessions.
- Ensure your reason aligns with the given persona characteristics.
- Do not include any additional comments or explanations outside the output you have to provide.
- Please make sure to give the output in the exact same format I provided
"""
\end{lstlisting}

\subsubsection{Item selection}

\par{Individual alignment}

\begin{lstlisting}[language=Python]
"""
Follow these instructions carefully:

1. You will be given some background information about a persona inside <background> tags. Use this information to execute the task.

<background>
{background}
</background>

2. You will be presented with a list of items inside <items> tags. These are the items available for purchase.

<items>
{items}
</items>

3. Your task is to choose one item from the list to buy and provide a reason for your choice, based on the background information you are provided with. Consider the following: a. Analyze the characteristics and infer the person's preferences and pickiness level. b. Evaluate each item in the list and how well it aligns with the background. c. Make a decision that best fits the persona's likely preferences and pickiness level that you need to infer.

4. Your response should be in valid JSON format, containing two keys: "title" (the product title) and "reason" (the explanation for your choice, including which is the level of pickiness you inferred). Be sure to use the same title you received as input in the <items> list.

5. Here's an example of the expected JSON output format:

<example>
{example_output}
</example>

6. Important notes:
- Only choose from the items provided in the list.
- Ensure your reason aligns with the given background information.
- Do not include any additional comments or explanations outside the JSON object.
- Make sure your JSON is properly formatted and valid.

7. Provide your response in a single JSON object, without any additional text before or after it.
"""
\end{lstlisting}

\par{Group alignment}
\begin{lstlisting}[language=Python]
"""
Follow these instructions carefully:

1. You will be given a set of persona characteristics inside <persona> tags. Embody this persona for the task.

<persona>
{persona}
</persona>

2. You will be presented with a list of items inside <items> tags. These are the items available for purchase.

<items>
{items}
</items>

3. Your task is to choose one item from the list to buy, based on the persona you are embodying. Consider the following: a. Analyze the persona characteristics and infer the person's preferences and pickiness level. b. Evaluate each item in the list and how well it aligns with the persona. c. Make a decision that best fits the persona's likely preferences and pickiness level that you need to infer.

4. Your response should be in valid JSON format, containing one key: "output", associated to a single value which is your answer. The value must be in valid integer format, which represents the index of the chosen item in the input list, (the indices start from 0).

5. Here's an example of the expected output in JSON format:

{example_output}

6. Important notes:
- Only choose from the items provided in the list.
- Ensure your reason aligns with the given persona characteristics.
- Do not include any additional comments or explanations outside the output integer.
- Please make sure to give the output in the exact same format I provided
"""
\end{lstlisting}

\subsubsection{Session generation}
\begin{lstlisting}[language=Python]
"""
You are presented with: (i) a detailed description of a customer 
of the online shopping website amazon.com and (ii) browsing tools. 
Your task is to impersonate the given customer, adhering to the provided 
description, and perform a shopping session.
You can browse the shopping website through the provided list of tools. 

Notes:
- You are submitting queries to the amazon.com search bar. Keep your queries simple and short (max 4 words). If you don't get search results, try up to 3 times making the query less specific each time.
- When you want to terminate the shopping session, remember to use the terminate_session tool.
- Before adding an item to the cart, you probably want to get more information on it by visiting the item's detail page through the get_product_info_tool.
- You can decide to buy the items added in the cart or not: If you are satisfied with the products added to the cart and believe the customer would proceed to a purchase, purchase the cart. Instead, if you think the customer might hesitate on the purchase, you can terminate the session without any purchases. 

Customer description: {persona}
"""
\end{lstlisting}

\section{KDE Bandwidth selection}
\label{appendix:bandwidth}
Below we report the KL divergence measurements for the query generation task for different choices of bandwidth.

\begin{table}[h]
\centering
\begin{tabular}{|l|c|c|c|} 
\hline 
\textbf{Bandwidth} & \textbf{Base} & \textbf{Base+Persona}\\
\hline 
0.001 & 410026 & \textbf{334170}\\
\hline
0.01 & 2236 & \textbf{1871}\\
\hline
0.1 & 18.81 & \textbf{17.51}\\
\hline
1 & 0.42 & \textbf{0.18}\\
\hline
\end{tabular}
\caption{KL divergence between human population and agents with/without persona for different values of KDE bandwidth.}
\end{table}

\end{document}